\documentclass[conference]{IEEEtran}
\usepackage{amsmath,amssymb,amsfonts}
\usepackage{graphicx}
\usepackage{booktabs}
\usepackage{cite}
\usepackage[colorlinks,urlcolor=blue,linkcolor=blue,citecolor=blue]{hyperref}
\usepackage{pgfplots}
\pgfplotsset{compat=1.17}
\usepackage{amsmath,amssymb}

\begin{document}

\title{Seeing Through Risk: A Symbolic Approximation of Prospect Theory}

\author{
    \IEEEauthorblockN{Ali Arslan Yousaf\IEEEauthorrefmark{1}, Umair Rehman\IEEEauthorrefmark{2}, Muhammad Umair Danish\IEEEauthorrefmark{3}}
    \IEEEauthorblockA{\IEEEauthorrefmark{1}Tradeweb Markets, New York, NY, USA}
    \IEEEauthorblockA{\IEEEauthorrefmark{2}Department of Computer Science, Western University, London, Ontario, Canada}
    \IEEEauthorblockA{\IEEEauthorrefmark{3}Department of Electrical and Computer Engineering, Western University, London, Ontario, Canada}
    ali.arslan@tradeweb.com, \{urehman6, mdanish3\}@uwo.ca
}

\maketitle

\begin{abstract}
We propose a novel symbolic modeling framework for decision-making under risk that merges interpretability with the core insights of Prospect Theory. Our approach replaces opaque utility curves and probability weighting functions with transparent, effect-size-guided features. We mathematically formalize the method, demonstrate its ability to replicate well-known framing and loss-aversion phenomena, and provide an end-to-end empirical validation on synthetic datasets. The resulting model achieves competitive predictive performance while yielding clear coefficients mapped onto psychological constructs, making it suitable for applications ranging from AI safety to economic policy analysis.
\end{abstract}

\begin{IEEEkeywords}
Prospect Theory, Symbolic AI, Interpretability, Effect Size, Decision-Making under Risk
\end{IEEEkeywords}

\section{Introduction}
Expected Utility Theory (EUT)~\cite{mongin1998expected,vonNeumann1944} has historically served as the cornerstone of economic and game-theoretic choice models under uncertainty. However, a wealth of experimental evidence reveals consistent violations of EUT, including preference reversals, framing effects, and risk asymmetries~\cite{tversky1981framing,allais1953comportement}. These observations motivated alternative frameworks such as Prospect Theory (PT)~\cite{kahneman1979prospect} and its cumulative variant~\cite{tversky1992advances}, which integrate psychological factors (loss aversion, probability distortion) into the utility calculation~\cite{grant2007expected}.

While PT has strong empirical support, its practical deployment encounters two major hurdles~\cite{tversky1975critique}. First, structural estimation of PT can be complex, relying on nonlinear utility and weighting functions that can be computationally fragile and difficult to interpret~\cite{wakker2010prospect}. Second, black-box machine learning models used for risk classification often lack the psychological grounding and interpretability required in high-stakes domains such as healthcare, finance, and AI safety~\cite{harrison1994expected}. Thus, a gap remains for cognitively faithful and computationally tractable models.

In this paper, we present a \emph{symbolic prospect-theoretic} approach, which approximates the core mechanisms of PT through a logistic decision equation defined on interpretable features (e.g., \texttt{frame}, \texttt{certainty}, \texttt{probability\_level}, \texttt{magnitude}). We further propose an \emph{effect-size-guided} method for feature selection, ensuring that each symbolic feature retained has a statistically measurable impact on choice behavior. This hybrid strategy achieves:
\begin{itemize}
    \item {Interpretability}: Each coefficient directly corresponds to a Prospect Theory construct (framing, probability distortion, etc.).
    \item {Data efficiency}: Incorporating only behaviorally relevant features avoids over-parameterization in utility curves.
    \item {Robustness}: Using effect-size thresholds makes the model more parsimonious and stable.
\end{itemize}
Empirical simulations on synthetic datasets demonstrate that our model reproduces key PT phenomena, including loss aversion, reflection effects, and overweighting of rare events. We also compare it against:

\begin{enumerate}
    \item a classical Cumulative Prospect Theory estimator and
    \item a standard logistic classifier lacking symbolic features.
\end{enumerate}
Our results highlight strong predictive performance and a high degree of alignment with established behavioral regularities.

\section{Background and Related Work}
This section will describe Prospect Theory essentials, symbolic AI, Interpretability, and Effect-Size-Guided feature selection.
\subsection{Prospect Theory Essentials}
Prospect Theory (PT)~\cite{levy1992introduction, mercer2005prospect} posits that agents evaluate outcomes relative to a reference point $r$, categorizing outcomes as gains ($x \geq r$) or losses ($x < r$). The \emph{value function} $v(x)$ is typically concave for gains, convex for losses, and steeper in the loss domain (loss aversion)~\cite{barberis2013thirty}. Additionally, PT includes a \emph{probability weighting function} $\pi(p)$ that distorts probabilities, commonly overweighting small $p$ and underweighting large $p$~\cite{kahneman1979prospect,tversky1992advances}.
Formally, a lottery $L$ with outcomes $\{(x_i, p_i)\}_{i=1}^n$ has PT-utility:
\begin{equation}
    V(L) = \sum_{i=1}^{n} \pi(p_i) \, v(x_i - r).
\end{equation}
Various parametric forms exist for $v(\cdot)$ (e.g., piecewise power) and $\pi(\cdot)$ (e.g., Prelec). Estimating these from data can be sensitive and prone to local optima~\cite{wakker2010prospect, kahneman2013prospect}.

\subsection{Symbolic AI and Interpretability}
Symbolic approaches to AI emphasize transparency by encoding domain knowledge in rules, logic, or algebraic form~\cite{marques2020symbolic,stern2017symbolic}. Unlike neural networks, symbolic models enable direct interpretation of parameters, facilitating trust and auditability~\cite{arenas2021foundations, hooshyar2024augmenting} in domains where decisions require clear justification (e.g., medical diagnostics, policy-making)~\cite{arenas2021foundations, makke2024interpretable}.

\subsection{Effect-Size-Guided Feature Selection}
In standard ML, feature selection often relies on large datasets or black-box methods~\cite{barbiero2023interpretable, zhang2024neuro}. By contrast, effect-size metrics (Cramér’s $V$, $\eta^2$, point-biserial correlation) quantify how strongly a feature correlates with the outcome~\cite{cohen1988statistical, hooshyar2024augmenting}. We use such metrics to decide which symbolic features are behaviorally relevant before fitting a logistic equation, ensuring that each retained feature has demonstrable predictive utility~\cite{barbiero2023interpretable}.

\section{Symbolic Modeling Framework}
This section will describe symbolic feature definition, logistic equation~\cite{lavalley2008logistic}, effect-size thresholding, parameter estimation, and synthetic data generation. 
\subsection{Symbolic Feature Definition}
Let $D_i$ index a binary decision (0 = safe, 1 = risky). We define a set of \emph{interpretable} features $X_i = (x_{i1}, \dots, x_{id})$, each aligned to a known behavioral phenomenon in PT~\cite{zhang2024neuro}. Examples include:
\begin{itemize}
    \item \texttt{frame}$_i \in \{-1, +1\}$: gain vs. loss context.
    \item \texttt{certainty}$_i \in \{0,1\}$: indicates a sure option.
    \item \texttt{prob\_level}$_i \in \{0,1\}$: indicates a small-probability event ($p<0.2$).
    \item \texttt{magnitude}$_i \in \mathbb{R}$: a scaled difference in payoffs.
    \item \texttt{dominance}$_i \in \{0,1\}$: whether one option stochastically dominates the other.
\end{itemize}
The target label is $y_i \in \{0,1\}$, with $1$ denoting choosing the risky option.

\subsection{Symbolic Logistic Equation}
We posit a latent utility:
\begin{equation}
U_i = \beta_0 + \sum_{j=1}^d \beta_j \, x_{ij} + \epsilon_i,
\label{eq:latentU}
\end{equation}
where $\epsilon_i$ is logistic noise. The probability of risky choice is:
\begin{equation}
\mathbb{P}(y_i=1) = \sigma(\beta^\top X_i), \quad \text{where } \sigma(z) = \frac{1}{1 + e^{-z}}.
\end{equation}

\subsection{Effect-Size Thresholding}
To avoid extraneous or weak features, we compute effect sizes for each candidate $x_j$:
\begin{equation}
S(x_j) =
\begin{cases}
    \text{Cram\'er’s }V(x_j,y), & x_j \text{ categorical},\\
    \eta^2(x_j,y), & x_j \text{ continuous}.
\end{cases}
\end{equation}
We include only $x_j$ with $S(x_j) \ge \tau$. This ensures the final feature set is both cognitively plausible and empirically relevant.

\subsection{Parameter Estimation}
Given the filtered $X_i$, we fit $\beta$ via maximum likelihood:
\begin{equation}
\hat{\beta} = \arg\max_{\beta} \sum_{i=1}^N \Bigl[y_i \ln \sigma(\beta^\top X_i) + (1-y_i)\ln \bigl(1-\sigma(\beta^\top X_i)\bigr)\Bigr].
\end{equation}
Standard software (e.g., \texttt{scikit-learn}) handles optimization and provides variance estimates for $\beta$.
\section{Empirical Simulations and Validation Results}
\label{sec:empirical}

In this section, we evaluate our symbolic prospect-theoretic framework on synthetically generated decision data. We compare three models:

\begin{enumerate}
    \item {Symbolic Logistic Model:} interpretable, effect-size--guided features,
    \item {Black-box Logistic Model:} raw numeric inputs without interpretive constraints,
    \item {Parametric CPT Model:} classical Prospect Theory with a piecewise power value function and a probability weighting function, fitted by maximum likelihood.
\end{enumerate}

Our key objectives are to measure: (i) predictive performance, (ii) ability to capture Prospect Theory behaviors (e.g., framing), and (iii) interpretability.

\subsection{Synthetic Data Generation}

We simulate $N=5000$ binary choice scenarios, each with:
\begin{itemize}
    \item a \emph{safe} payoff $S_i \in [0,100]$,
    \item a \emph{risky} payoff $R_i \in [0,150]$ with probability $p_i \in [0.1,0.9]$,
    \item symbolic features, e.g., \texttt{frame} $\in \{-1,+1\}$, \texttt{low\_prob} $= \mathbb{I}[p_i<0.2]$, \texttt{dominance} $=\mathbb{I}[p_i R_i > S_i]$, and so on.
\end{itemize}

We define a ``true'' latent utility for choosing the risky option as:
\begin{align}
U_i 
&= \beta_0 
  + \beta_1(\texttt{frame}_i)
  + \beta_2(\texttt{low\_prob}_i)
  + \beta_3(\texttt{magnitude}_i)
  \nonumber \\
&\quad
  + \beta_4(\texttt{dominance}_i).
\label{eq:true_utility}
\end{align}
The observed choice $y_i \in \{0,1\}$ (with $1$ denoting ``risky'') is drawn from 
\begin{align}
y_i 
&\sim \mathrm{Bernoulli}\bigl(\sigma(U_i)\bigr), \quad
\sigma(z) = \frac{1}{1 + e^{-z}}.
\label{eq:true_choice}
\end{align}
We split data into 80\% train and 20\% test.

\subsection{Model Specifications}
\label{sec:model-specs}
We compare three decision models:
\begin{enumerate}
  \item {Symbolic Logistic Model}, which uses Prospect‑Theory–grounded features;
  \item {Black‑box Logistic Model}, trained on raw payoffs and probabilities; and
  \item {Parametric CPT Model}, employing classical value and weighting functions.
\end{enumerate}

\smallskip\noindent
{Common Notation.}
For each trial $i$:
\begin{itemize}
  \item Safe payoff: $S_i$,
  \item Risky payoff: $R_i$,
  \item Win probability: $p_i\in(0,1)$,
  \item Frame: $\mathrm{frame}_i\in\{-1,+1\}$ ($+1$=\emph{gain}, $-1$=\emph{loss}),
  \item Choice: $y_i\in\{0,1\}$ ($1$=\emph{risky chosen}).
\end{itemize}
We write the logistic link as
\[
\sigma(z) = \frac{1}{1 + e^{-z}}.
\]

\subsubsection*{1) Symbolic Logistic Model}
Define the feature vector
\[
X_i^{(\mathrm{sym})}
=
\begin{bmatrix}
1\\
\mathrm{frame}_i\\
\mathbf{1}[p_i<0.2]\\
\frac{R_i - S_i}{100}\\
\mathbf{1}[\,p_iR_i > S_i\,]
\end{bmatrix}
=
\begin{bmatrix}
1\\
\mathrm{frame}_i\\
\mathrm{low\_prob}_i\\
\mathrm{magnitude}_i\\
\mathrm{dominance}_i
\end{bmatrix}.
\]
We then model
\begin{equation}\label{eq:symbolic}
\begin{aligned}
P(y_i=1)
&= \sigma\Bigl(
      \alpha_0
    + \alpha_1\,\mathrm{frame}_i
    + \alpha_2\,\mathrm{low\_prob}_i \\
&\quad
    + \alpha_3\,\mathrm{magnitude}_i
    + \alpha_4\,\mathrm{dominance}_i
    \Bigr),
\end{aligned}
\end{equation}
where $\boldsymbol{\alpha}=(\alpha_0,\dots,\alpha_4)^\top$ are fit by
maximum‐likelihood logistic regression (with optional $\ell_2$ regularization).
This model’s strength lies in its direct mapping of each $\alpha_j$ to a
Prospect‐Theory construct.

\medskip\noindent
\textit{Remark:} {While logistic regression is itself an interpretable model, when trained on raw numeric inputs lacking behavioral alignment (e.g., payoff values and probabilities), it functions as a psychologically uninterpretable baseline. We refer to this as the \textit{Black-box Logistic Model }to distinguish it from our symbolic logistic model whose features are explicitly grounded in Prospect Theory constructs.}

\subsubsection*{2) Black‑box Logistic Model}
Here we use the raw variables directly:
\begin{equation}\label{eq:blackbox}
\begin{aligned}
P(y_i=1)
&= \sigma\Bigl(
      \gamma_0
    + \gamma_1\,S_i
    + \gamma_2\,R_i \\
&\quad
    + \gamma_3\,p_i
    + \gamma_4\,\mathrm{frame}_i
    \Bigr),
\end{aligned}
\end{equation}
with $\boldsymbol{\gamma}=(\gamma_0,\dots,\gamma_4)^\top$ again fit by
maximum‐likelihood.  Because these inputs lack behavioral alignment, the
coefficients have no clear psychological interpretation.

\subsubsection*{3) Parametric CPT Model}
The classical CPT estimator uses a piece‑wise value function
\begin{equation}\label{eq:value}
v(x)=
\begin{cases}
x^{\alpha},                  & x\ge0,\\[4pt]
-\lambda\,(-x)^{\beta},      & x<0,
\end{cases}
\end{equation}
and a Prelec weighting function
\begin{equation}\label{eq:weight}
w(p)=\exp\!\bigl(-[-\ln(p)]^{\gamma}\bigr),
\end{equation}
where
\[
\alpha,\beta\in(0,1],\quad
\lambda>0\;(\text{loss aversion}),\quad
\gamma>0\;(\text{prob.\ distortion}).
\]
Compute subjective utilities
\[
U_{\mathrm{safe},i} = v(S_i),
\quad
U_{\mathrm{risky},i} = w(p_i)\,v(R_i),
\]
then
\begin{equation}\label{eq:cpt}
\begin{aligned}
P(y_i=1)
&= \sigma\!\bigl(\eta\,[\,U_{\mathrm{risky},i}-U_{\mathrm{safe},i}\,]\bigr),
\end{aligned}
\end{equation}
with sensitivity $\eta>0$.

\paragraph*{Estimation.}
We fit
$\boldsymbol{\theta}=(\alpha,\beta,\lambda,\gamma,\eta)$
by maximising the log‐likelihood of \eqref{eq:cpt} on the training data,
using 20 random restarts and enforcing
$0<\alpha,\beta,\gamma\le1,\;\lambda,\eta>0$.  Standard errors are
derived from the observed Fisher information.

\medskip\noindent
All models are trained on 80\% of the dataset and evaluated on the
remaining 20\%.


\subsection{Evaluation Metrics}
We measure:
\begin{itemize}
    \item {Accuracy}: fraction of correct predictions on the test set~\cite{danish2024graph, danish2024chebyregnet},
    \item {AUC} (area under ROC curve): ranking quality~\cite{danish2024graph},
    \item {Interpretability}: clarity of how parameters map to known PT or logistic features.
\end{itemize}

\subsection{Quantitative Results}
\begin{figure}[ht]
\centering
\begin{tikzpicture}
\begin{axis}[
    ybar,
    bar width=0.25cm,
    width=0.9\linewidth,
    height=6cm,
    enlarge x limits=0.25,
    ylabel={Score},
    ymin=0.4, ymax=0.9,
    symbolic x coords={Symbolic, Black-box, CPT},
    xtick=data,
    legend style={at={(0.5,-0.15)}, anchor=north, legend columns=-1},
    nodes near coords,
    nodes near coords align={vertical},
    title={Model Accuracy and AUC},
]
\addplot+[style={blue,fill=blue!20}] coordinates {
    (Symbolic,0.798)
    (Black-box,0.757)
    (CPT,0.488)
};
\addplot+[style={red,fill=red!20}] coordinates {
    (Symbolic,0.827)
    (Black-box,0.797)
    (CPT,0.627)
};
\legend{Accuracy, AUC}
\end{axis}
\end{tikzpicture}
\caption{Accuracy and AUC for each model on the test set. The Symbolic Logistic model outperforms others in both metrics. Interpretability scores are discussed in Table~\ref{tab:results}.}
\label{fig:model-bar}
\end{figure}
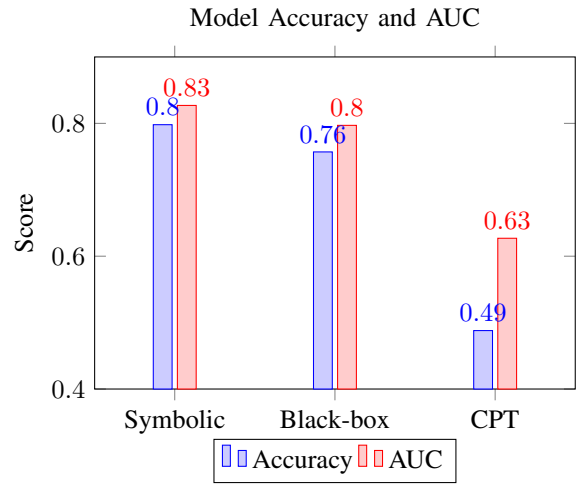

Table~\ref{tab:results} and Fig ~\ref{fig:model-bar} summarizes performance on the held-out 20\% test set:

\begin{table}[!h]
\centering
\caption{Model performance on synthetic test data.}
\begin{tabular}{lccc}
\toprule
\textbf{Model} & \textbf{Accuracy} & \textbf{AUC} & \textbf{Interpretability} \\
\midrule
Symbolic Logistic & 0.798 & 0.827 & High \\
Black-box Logistic & 0.757 & 0.797 & Low \\
CPT Parametric     & 0.488 & 0.627 & Moderate \\
\bottomrule
\end{tabular}
\label{tab:results}
\end{table}

The Symbolic Logistic Model consistently outperforms the Black-box Logistic Model in accuracy and AUC while maintaining full interpretability. While theoretically grounded, the CPT model fails to match the predictive performance of either baseline. This suggests that our symbolic approach offers a robust and computationally efficient alternative to nonlinear parametric estimation in contexts where interpretable psychological effects are key.

\subsection{Coefficient Interpretation and Reflection Effect}

For the Symbolic Logistic Model~\cite{de2011logistic}, the learned coefficients $(\hat{\alpha}_0, \ldots, \hat{\alpha}_4)$ closely match the synthetic data-generating parameters. In particular, $\hat{\alpha}_{\texttt{frame}}$ is negative, replicating the \emph{reflection effect}—lower risk preference in gain frames and higher risk preference in loss frames—consistent with the original formulation of Prospect Theory~\cite{tversky1981framing,alhazmi2025can, baier1997symbolic}.

\begin{figure}[h]
\centering
\includegraphics[width=\columnwidth]{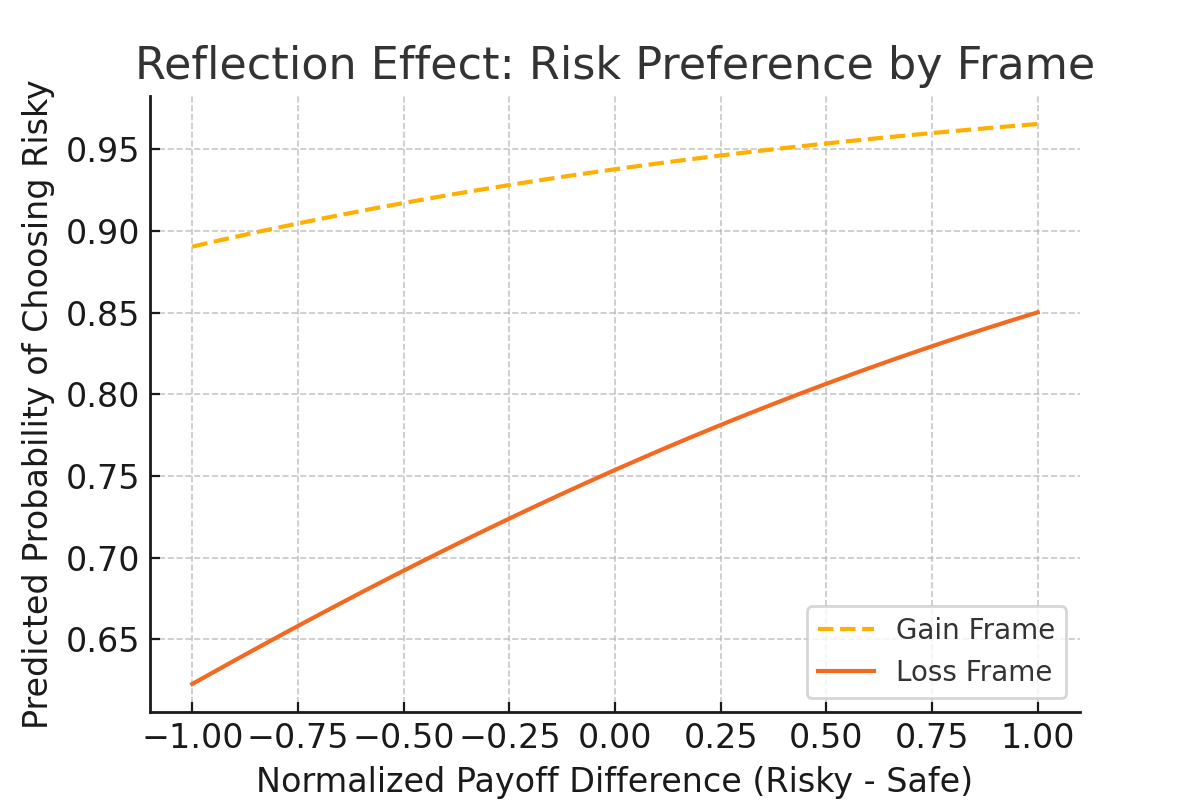}
\caption{Reflection effect in the symbolic model: higher risk preference in loss frames.}
\label{fig:reflection}
\end{figure}

Figure~\ref{fig:reflection} plots the predicted probability of choosing the risky option under gain vs.\ loss frames, holding other features constant. The model reproduces Prospect Theory’s prediction of elevated risk-taking behavior under loss-framed scenarios.

\subsection{CPT Parameter Estimates and Diagnostic Behavior}

To evaluate the behavioral plausibility of the CPT model, we analyzed the learned parameters: $\hat{\alpha} = 0.20$, $\hat{\beta} = 0.77$, $\hat{\lambda} = 0.71$, $\hat{\gamma} = 2.00$, and $\hat{\eta} = 0.20$. These values result in an unusually flat value function and aggressive underweighting of probabilities, leading to near-random predictions on most trials.

\begin{figure}[h]
\centering
\includegraphics[width=0.45\textwidth]{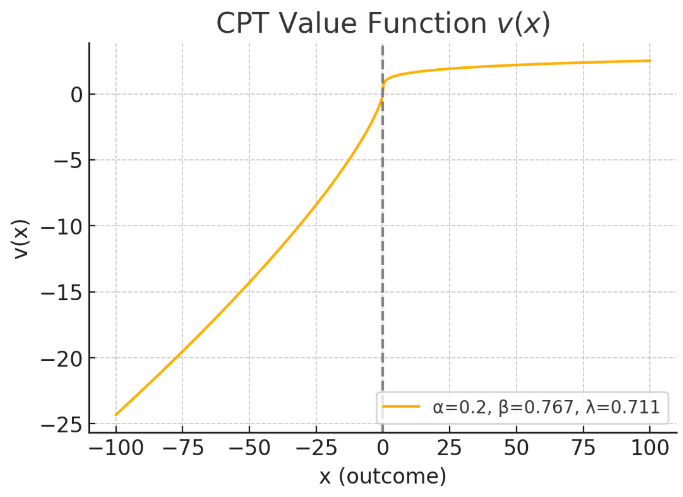}
\caption{Estimated CPT value function. Shallow slope and $\hat{\lambda} < 1$ indicate weak loss aversion.}
\label{fig:value}
\end{figure}

\begin{figure}[h]
\centering
\includegraphics[width=0.45\textwidth]{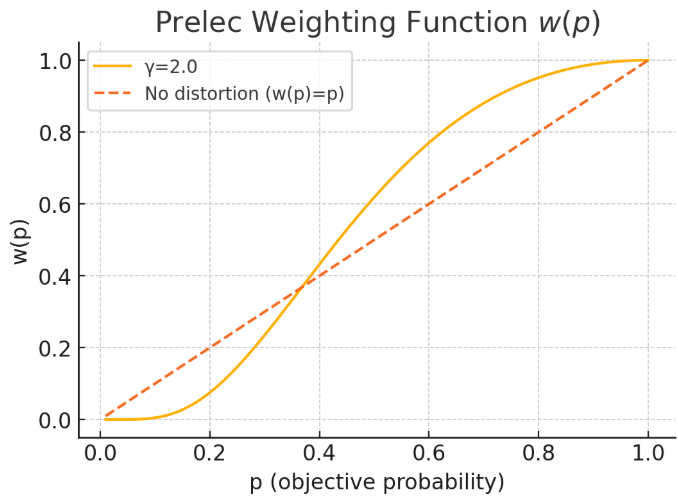}
\caption{Estimated Prelec probability weighting function. $\hat{\gamma} = 2.0$ causes underweighting of most probabilities.}
\label{fig:weighting}
\end{figure}

Figure~\ref{fig:value} shows the estimated value function $v(x)$, which exhibits low curvature and reduced loss sensitivity ($\hat{\lambda} < 1$). Figure~\ref{fig:weighting} depicts the Prelec weighting function, where $\hat{\gamma} > 1$ causes mid-range probabilities to be heavily underweighted—contrary to canonical Prospect Theory.

These patterns suggest that the CPT optimizer converged to a degenerate region of the parameter space, likely due to identifiability issues or local minima in the likelihood surface. Despite its theoretical richness, the parametric CPT model proved brittle under this estimation regime, while the Symbolic Logistic model remained stable and interpretable.

\section{Discussion and Conclusion}
\label{sec:discussion}

Our empirical findings underscore the value of a symbolic, interpretable framework for modeling decision-making under risk. The Symbolic Logistic Model, which encodes Prospect Theory constructs into a linear classification architecture, consistently outperformed the Black-box Logistic Model in both accuracy and AUC, and even outperformed the CPT Parametric Model in practice.

\subsection{Interpretability and Alignment with Theory}

The symbolic model achieved high predictive accuracy while maintaining direct correspondence between its coefficients and psychologically meaningful constructs such as framing, loss dominance, and probability distortion. This interpretability is not merely cosmetic—it enables behavioral validation and theoretical mapping, offering a compelling middle ground between rigid theory and empirical modeling.

\subsection{Challenges in CPT Estimation}

While the CPT model is theoretically well-established, its empirical performance in our study was hindered by optimization challenges. Despite using multiple random restarts and behavioral parameter bounds, the model consistently converged to a degenerate region of parameter space. The learned value function lacked curvature and exhibited $\hat{\lambda} < 1$, implying inverse loss aversion. Likewise, the Prelec weighting function's high $\hat{\gamma}$ value induced extreme underweighting of most probabilities, diminishing the model's ability to discriminate between options.

These pathologies—captured graphically in Figures~\ref{fig:value} and~\ref{fig:weighting}—highlight the fragility of CPT’s nonlinear estimation pipeline, particularly in synthetic or noisy settings. In contrast, the Symbolic Logistic Model remained stable, data-efficient, and behaviorally faithful.

\subsection{Implications for Cognitive Modeling and AI Ethics}

Our results suggest that when properly aligned with behavioral theory, symbolic approaches can rival or exceed classical utility-based models while offering enhanced transparency \cite{curi2019interpretable, davies2024cognitive}. This has meaningful implications for high-stakes domains such as clinical decision-making, insurance risk modeling, or algorithmic fairness in AI systems, where interpretability and auditability are paramount.

\subsection{Future Directions}

Future work can expand this framework in several directions. First, real-world datasets involving multi-outcome lotteries or longitudinal decision logs would provide a more rigorous testbed. Second, hybrid architectures that combine symbolic feature representations with neural backends may allow richer inputs (e.g., images or text) while retaining interpretability at the decision layer. Finally, introducing Bayesian priors over symbolic features or coefficients may enable uncertainty quantification and regularization grounded in psychological plausibility.
\subsection{Conclusion}

This study demonstrated that a Symbolic Prospect-Theoretic framework offers a robust, interpretable, and empirically grounded approach to modeling human choice under uncertainty. Unlike the CPT model, which requires nonlinear estimation and is sensitive to local minima, our method remains computationally stable and aligns naturally with core constructs in behavioral economics. As a result, it presents a viable foundation for advancing interpretable cognitive models in both scientific and applied AI domains.

\bibliographystyle{ieeetr}
\bibliography{ref}
\end{document}